\documentclass[letterpaper, 10 pt, conference]{ieeeconf}  

\IEEEoverridecommandlockouts                              

\usepackage{graphics} 
\usepackage{epsfig} 
\usepackage{amsmath} 
\usepackage{amssymb}  
\usepackage{graphicx}

\title{\LARGE \bf
Multi-Agent Reinforcement Learning and Real-Time Decision-Making in Robotic Soccer for Virtual Environments
}

\author{Aya Taourirte\textsuperscript{1} and Md Sohag Mia\textsuperscript{1,$\dagger$}
\thanks{$\dagger$ Corresponding author}
\thanks{\textsuperscript{1}Nanjing University of Information Science and Technology, Nanjing, China}
}

\begin{document}

\maketitle
\thispagestyle{empty}
\pagestyle{empty}

\begin{abstract}

The deployment of multi-agent systems in dynamic, adversarial environments like robotic soccer necessitates real-time decision-making, sophisticated cooperation, and scalable algorithms to avoid the curse of dimensionality. While Reinforcement Learning (RL) offers a promising framework, existing methods often struggle with the multi-granularity of tasks (long-term strategy vs. instant actions) and the complexity of large-scale agent interactions. This paper presents a unified Multi-Agent Reinforcement Learning (MARL) framework that addresses these challenges. First, we establish a baseline using Proximal Policy Optimization (PPO) within a client-server architecture for real-time action scheduling, with PPO demonstrating superior performance (4.32 avg. goals, 82.9\% ball control). Second, we introduce a Hierarchical RL (HRL) structure based on the options framework to decompose the problem into a high-level trajectory planning layer (modeled as a Semi-Markov Decision Process) and a low-level action execution layer, improving global strategy (avg. goals increased to 5.26). Finally, to ensure scalability, we integrate mean-field theory into the HRL framework, simplifying many-agent interactions into a single agent vs. the population average. Our mean-field actor-critic method achieves a significant performance boost (5.93 avg. goals, 89.1\% ball control, 92.3\% passing accuracy) and enhanced training stability. Extensive simulations of 4v4 matches in the Webots environment validate our approach, demonstrating its potential for robust, scalable, and cooperative behavior in complex multi-agent domains.

\end{abstract}

\section{Introduction}

With the rapid advancement of electronic, computer, communication, and artificial intelligence technologies, the scope of robotics applications has expanded significantly \cite{1,2,3,4}. Among the most widely adopted research platforms is the NAO humanoid robot, developed by Aldebaran Robotics, which integrates sensing, vision, and control in a compact design \cite{5,6}. NAO robots have become a standard tool in robotic soccer research, providing an effective testbed for studying multi-agent collaboration, planning, and decision-making in dynamic environments \cite{7,8,9}. 

The concept of robotic soccer was introduced in the 1990s by Mackworth \cite{10}, leading to the establishment of RoboCup, which has since grown into the largest international competition for autonomous robots \cite{11,12}. RoboCup’s long-term vision is to build a team of fully autonomous humanoid robots capable of defeating the FIFA World Cup champions by 2050 \cite{13}. This ambitious target has made robotic soccer a benchmark domain for multi-agent systems (MAS), combining real-time perception, strategic decision-making, and cooperative behaviors.

Robotic soccer is inherently a multi-agent problem where robots must learn from continuous interaction with the environment and make optimal decisions under uncertainty. Reinforcement Learning (RL) provides a powerful framework for this, as it enables agents to optimize long-term cumulative rewards through trial-and-error learning. However, robotic soccer also introduces unique challenges: tasks span multiple temporal scales (e.g., long-term trajectory planning versus short-term motor actions), and effective strategies require coordination among several agents. Existing methods, often based on either offline or online planning \cite{14,15}, struggle to adapt to the fast-changing dynamics of the field and to fully exploit the robots’ capabilities for real-time execution and cooperation. This creates an urgent need for methods that unify real-time decision-making, multi-granularity task decomposition, and scalable multi-agent coordination.

\begin{figure}[t]
\centering
\includegraphics[width=0.48\textwidth]{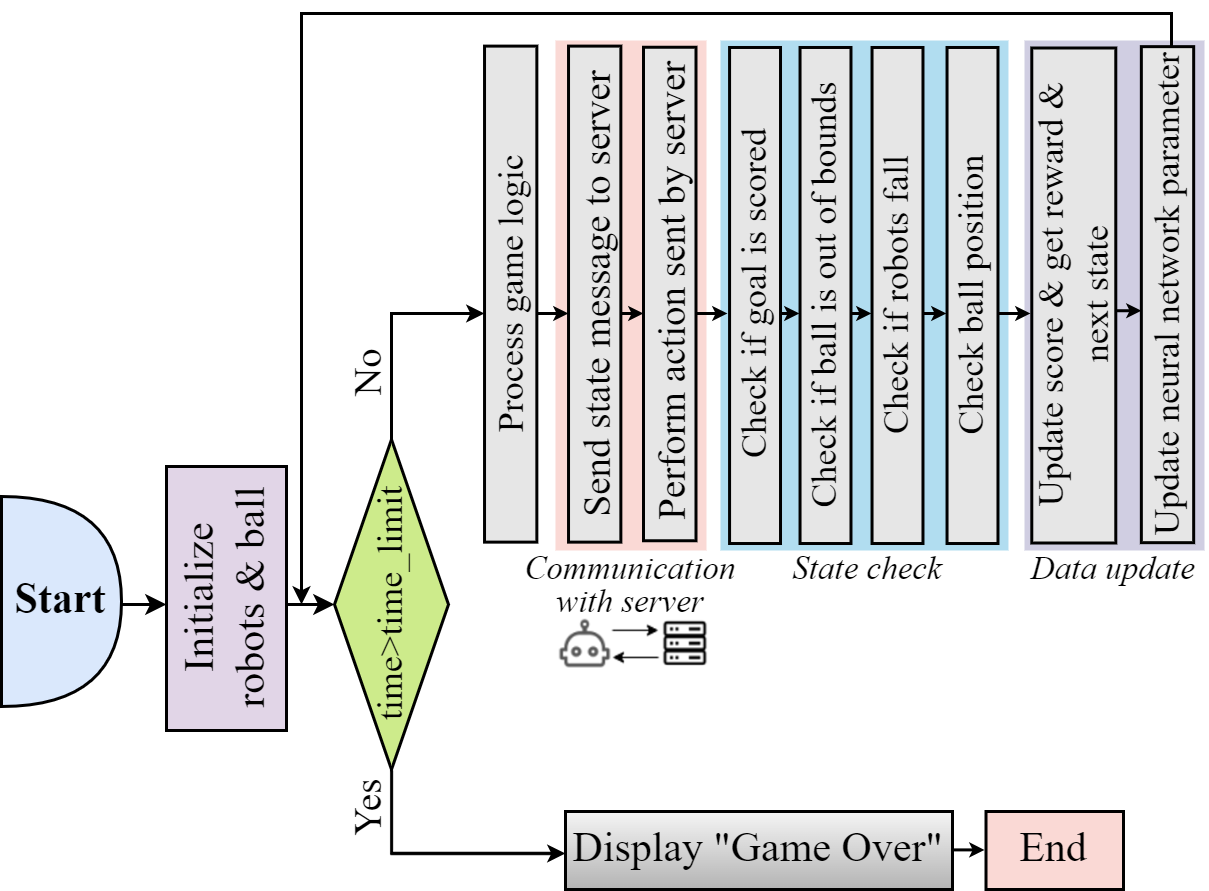}
\caption{Client-Server Match Control Logic Flow. The diagram illustrates the end-to-end process from initialization and state checking to server communication, action execution, goal detection, and neural network updates for real-time robotic soccer decision-making.}
\label{fig:logic_control}
\end{figure}

In this work, we address these challenges by proposing a unified Multi-Agent Reinforcement Learning (MARL) framework for robotic soccer. First, we formulate the robot action scheduling problem as a Markov Decision Process (MDP) and establish baselines using Proximal Policy Optimization (PPO). PPO demonstrates clear advantages in stability and performance for real-time control. Second, to handle multi-task and multi-granularity decisions, we introduce a Hierarchical Reinforcement Learning (HRL) model based on the options framework, where high-level trajectory planning operates on a coarse timescale while low-level control manages fine-grained actions. Finally, to ensure scalability to larger teams, we integrate mean-field theory into the hierarchical framework, transforming many-to-many agent interactions into tractable mean-field approximations. This hierarchical mean-field actor-critic method achieves both higher performance and improved training stability in large-scale multi-agent environments.

The proposed framework is validated through extensive simulations in the Webots platform using 4v4 NAO soccer matches. Results demonstrate clear progression in performance: PPO surpasses DQN in convergence and stability, HRL enhances global strategy via trajectory planning, and the mean-field extension achieves the best overall results, with significant improvements in goals scored, ball control, and passing accuracy. These findings highlight the potential of combining hierarchical reinforcement learning and mean-field theory to advance multi-agent decision-making in robotic soccer and beyond. As shown in Fig.~\ref{fig:logic_control}, the overall match logic control flow is illustrated.

\section{Preliminaries}

The NAO humanoid robot has become a standard platform for RoboCup competitions and academic research, providing a unified hardware and software environment for testing decision-making and multi-agent coordination \cite{16}. Since its adoption in 2007 as the official robot of the Standard Platform League, NAO has been extensively studied in areas such as locomotion, motion planning, and perception. For instance, Kashyap et al. developed whole-body control methods for stable humanoid walking, while Xi et al. proposed hybrid reinforcement learning strategies to enhance walking robustness on static and dynamic surfaces \cite{17}. In motion planning, researchers have introduced algorithms such as SMHA* for whole-body footprint planning \cite{18} and Dynamic Window Approach (DWA) variants optimized with metaheuristics for real-time path adjustment \cite{19}. In terms of perception and localization, visual odometry and SLAM-based approaches have been adapted for NAO, including dense RGB-D methods robust to dynamic scenes \cite{20} and CNN-based visual map localization \cite{21}. These studies highlight the NAO robot’s role as a versatile testbed for advancing control, perception, and decision-making algorithms.

Trajectory planning remains central to robot soccer, where stability and efficiency are crucial. Research spans both global and local path planning methods. Classical search-based algorithms such as Dijkstra and A* \cite{25,26,27} have been extended for multi-goal scenarios and optimized with graph methods, while sampling-based approaches like RRT and PRM address high-dimensional planning \cite{30,31,32}. Bio-inspired optimization techniques, including genetic algorithms and particle swarm optimization \cite{33,34}, further enhance adaptability in dynamic environments. Local planning methods such as the Dynamic Window Approach (DWA) \cite{35,36} and its learning-based variants \cite{37,39,40,41}, as well as Time Elastic Band (TEB) \cite{42}, are widely applied to enable robots to adapt to rapidly changing environments, making them especially suitable for RoboCup. The complementary nature of global and local planning ensures both strategic foresight and real-time reactivity in soccer matches.

Beyond locomotion and planning, decision-making systems have evolved significantly. Earlier works focused on opponent strategy recognition \cite{43} and role allocation mechanisms \cite{44}. More recently, reinforcement learning has been applied to robotic soccer, enabling teams to autonomously develop cooperative strategies under dynamic conditions \cite{45}. These advances demonstrate the growing importance of learning-based methods for handling complex multi-agent interactions in real time.

All experiments in this work are conducted in Webots, a widely used simulator that provides accurate modeling of NAO robots, including kinematics, dynamics, and sensor feedback \cite{46,48}. Compared to other platforms such as Gazebo, V-REP, and MuJoCo, Webots offers strong support for humanoid robots, real-time visualization, and seamless integration with reinforcement learning algorithms, making it particularly well-suited for robotic soccer research.

In summary, prior work has addressed various aspects of locomotion, planning, perception, and decision-making in robotic soccer. However, existing methods often lack the ability to simultaneously manage multi-granularity decision-making and scalable multi-agent interactions in dynamic settings. This motivates the development of our unified reinforcement learning framework, which integrates hierarchical control and mean-field theory for efficient and cooperative robotic soccer strategies.

\section{Methodology}

This section presents our integrated MARL framework designed to address real-time decision-making, strategic hierarchy, and multi-agent scalability. We first introduce a PPO-based action scheduler for real-time control, then extend it with a hierarchical RL architecture to separate high-level planning from low-level execution. Finally, we incorporate mean-field theory to enable efficient coordination among multiple agents, ensuring both performance and scalability in dynamic environments.

\subsection{Action Scheduling Optimization with PPO}

To address the challenge of real-time decision-making in the highly dynamic environment of a robotic soccer match, we formulate the robot action scheduling problem as a Markov Decision Process (MDP). The core objective is to learn a policy $\pi$ that maps states $\mathbf{s}_t$ to actions $\mathbf{a}_t$ to maximize the discounted long-term reward $\bar{R} = \mathbb{E}\left[\sum_{t=0}^{\infty} \gamma^t r_t\right]$.

The state observation for each agent (robot) is designed for effective decision-making and is defined as:
\begin{equation}
\mathbf{s}_t = [P, P_{\text{goal}}, if_c, P^{\text{oppo}}] \in \mathcal{S},
\end{equation}
where $P$ and $P^{\text{oppo}}$ are the positions of teammates and opponents, $P_{\text{goal}}$ is the ball position, and $if_c$ is a binary vector indicating which robot has control of the ball. The immediate reward $r_t$ is a composite function designed to incentivize scoring goals, moving towards the ball, and maintaining control:
\begin{equation}
r_t = \frac{1}{-\log\left(r_{t}^{\text{goal}} + r_{t}^{\text{close}} + r_{t}^{\text{control}}\right)}.
\end{equation}
The action space for each robot is discrete, comprising actions such as \textit{move forward}, \textit{turn left}, \textit{turn right}, \textit{shoot}, and \textit{stay}.

A Client-Server (C-S) architecture centralizes computation for multi-agent coordination: robots (clients) transmit local observations to a server, which maintains the global state, runs the PPO algorithm to compute optimal actions, and dispatches commands for synchronized execution.

We employ a Proximal Policy Optimization (PPO) algorithm based on an Actor-Critic (AC) architecture for policy optimization. The actor ($\pi_\theta(a_t | s_t)$) selects actions, while the critic ($V_w(s_t)$) estimates state value and reduces policy gradient variance. PPO enhances stability via a clipped surrogate objective that prevents large policy updates. The algorithm proceeds by: collecting trajectories $(s_t, a_t, r_t, s_{t+1})$ under the current policy, computing discounted returns $R_t$, and normalizing them for training stability:
\begin{equation}
R_t^{\text{norm}} = \frac{R_t - \mu}{\sigma}.
\end{equation}

The advantage function $A_t$, which measures how much better an action is than average, is then estimated using the critic's value function:
\begin{equation}
A_t = R_t^{\text{norm}} - V_w(s_t).
\end{equation}

The core of PPO is its objective function. The probability ratio between the new and old policies is calculated as:
\begin{equation}
r_t(\theta) = \frac{\pi_\theta(a_t | s_t)}{\pi_{\theta_{\text{old}}}(a_t | s_t)}.
\end{equation}

This ratio is used in a clipped surrogate objective function to constrain the policy update:
\begin{equation}
L^{\text{CLIP}}(\theta) = \mathbb{E}_t\left[ \min\left( r_t(\theta) A_t, \text{clip}(r_t(\theta), 1-\epsilon, 1+\epsilon) A_t \right) \right],
\end{equation}

where $\epsilon$ is a hyperparameter (e.g., 0.2) that defines the clipping range. This clipping prevents the policy from changing too radically in a single update. The total loss function minimized by PPO combines this clipped objective with a value function error term and an entropy bonus for exploration:
\begin{equation}
L(\theta, w) = L^{\text{CLIP}}(\theta) + c_1 L^V(w) - c_2 H(\pi_\theta),
\end{equation}
where $L^V(w) = \mathbb{E}_t[(V_w(s_t) - R_t^{\text{norm}})^2]$ is the value loss, $H(\pi_\theta)$ is the policy entropy, and $c_1$, $c_2$ are weighting coefficients.

This formulation allows our robotic agents to learn a performant and stable action-scheduling policy directly from sensor data and rewards, enabling real-time decision-making capable of adapting to the unpredictable flow of a soccer match.

\subsection{Hierarchical RL for Multi-Task Decisions}

\begin{figure}[t]
\centering
\includegraphics[width=0.5\textwidth]{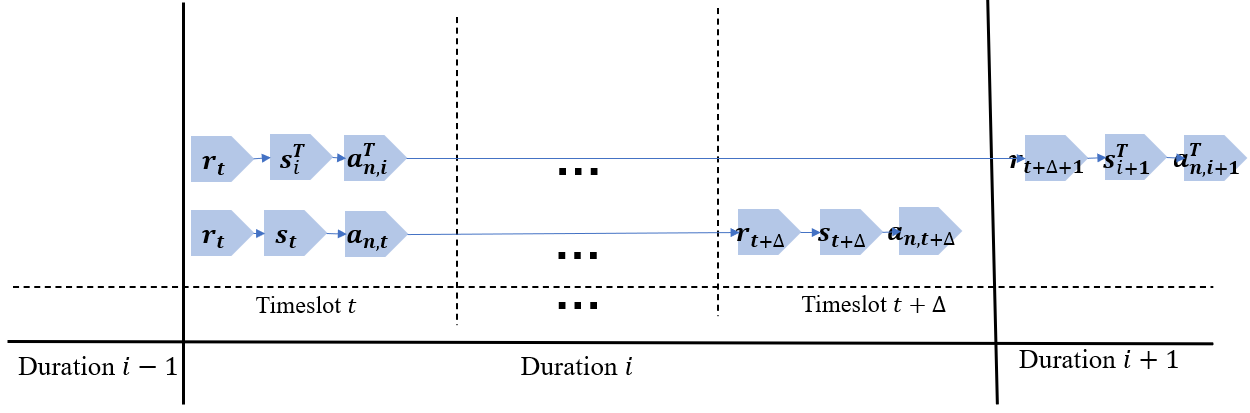}
\caption{Hierarchical decision-making structure.}
\label{fig:HRL}
\end{figure}

To optimize global team strategy, we extend the real-time action scheduler with coarse-grained trajectory planning. This introduces a multi-task, multi-granularity decision problem: high-level trajectory planning operates on a longer timescale, defining strategic goals over $\Delta$ timesteps, while low-level action scheduling requires real-time, fine-grained motor control. The different temporal abstractions and objectives of these tasks make a flat MDP formulation inefficient.

We address this via a Hierarchical Reinforcement Learning (HRL) framework based on the options formalism and Semi-Markov Decision Processes (SMDPs). An option $o = \langle \mathcal{I}^o, \pi^o, \beta^o \rangle$ represents a temporally extended action, where $\mathcal{I}^o \subseteq \mathcal{S}$ is the initiation set, $\pi^o$ is the option's policy, and $\beta^o$ is the termination condition. This allows us to model the high-level task as an SMDP $(\mathcal{S}, \mathcal{O}, \mathcal{P}^o, \mathcal{R}^o, \gamma)$, where options $\mathcal{O}$ are the actions.

Our hierarchical structure, depicted in Fig.~\ref{fig:HRL}, consists of two layers: (a) High-Level (Trajectory Planning): Modeled as an SMDP, this layer selects a directional option $a_{n,i}^T \in \{1, \ldots, 8\}$ every $\Delta$ timesteps for each robot $n$, defining a coarse movement trajectory. The joint option is $\mathbf{a}_i^T = [a_{1,i}^T, \ldots, a_{N,i}^T]$. (b) Low-Level (Action Scheduling): Modeled as an MDP, this layer executes the primitive actions (e.g., move, kick, turn) in each timestep $t$ to implement the high-level option, governed by the policy $a_{n,t} \sim \pi_n(\cdot | \mathbf{s}_t; \theta_n)$.

The state observation for the high-level planner is $\mathbf{s}_{i}^T = [P, P_{\text{goal}}, if_c, P^{\text{oppo}}]$, which is shared with the low layer. The high-level option $a_{n,i}^T$ directly influences the robot's position, which in turn becomes part of the state input for the low-level policy, creating a tight coupling between the two layers.

A key innovation is the design of a reward function that explicitly promotes multi-robot cooperation and strategic play:
\begin{align}
    r_t^{\text{goal}} &= 1  \\
    r_t^{\text{close}} &= \sum_{n=1}^{N} 0.1 \frac{1}{d_n}  \\
    r_t^{\text{control}} &= r^{\text{control}} + \sum_{k \in \mathcal{N}(n)} 0.1 \frac{1}{d_k^{\text{goal}}} 
\end{align}

where $d_n$ is the distance to the ball, $d_k^{\text{goal}}$ is the distance of non-ball-holder $k$ to the opponent's goal, OF(offensive positioning) and $r^{\text{control}}=0.4$. This reward structure incentivizes robots to pressure the ball, position themselves advantageously for shots, and ultimately score.

We employ an Actor-Critic architecture for both layers. The high-level policy $\pi_n^T(\cdot | \mathbf{s}_i^T; \theta_n^T)$ selects options. Its value function $V_n^T(\mathbf{s}_i^T; w_n^T)$ is updated using the TD error computed over the $\Delta$-step duration:
\begin{equation}
    \delta_{n,i}^T = r_{i+\Delta} + \gamma V_n^T(\mathbf{s}_{i+1}^T; w_n^T) - V_n^T(\mathbf{s}_{i}^T; w_n^T).
\end{equation}
The critic and actor parameters are then updated as:
\begin{align}
    w_n^T &\leftarrow w_n^T + \beta \delta_{n,i}^T \nabla_{w_n^T} V_n^T(\mathbf{s}_{i}^T; w_n^T), \\
    \theta_n^T &\leftarrow \theta_n^T + \alpha \delta_{n,i}^T \nabla_{\theta_n^T} \log \pi_n^T(\mathbf{s}_{i}^T; \theta_n^T).
\end{align}

The low-level policy $\pi_n(\cdot | \mathbf{s}_t; \theta_n)$ for action execution is updated at every timestep using the standard PPO objective, guided by the reward $r_t$ and the TD error $\delta_{n,t}$.

This hierarchical decomposition enables efficient learning of complex strategies by separating long-term planning from short-term execution, allowing the robots to dynamically coordinate their offensive and defensive formations.

\subsection{MDO based on Mean-field Theory}

The hierarchical framework introduces multiple homogeneous agents per layer, causing the joint state-action space to grow exponentially with the number of robots—a phenomenon known as the curse of dimensionality. To enable scalable coordination among homogeneous agents within each layer, we integrate mean-field theory into our hierarchical RL framework.

We model the fully cooperative multi-agent trajectory planning problem as a stochastic game $\Gamma \triangleq (S, A_1, \cdots, A_N, r, P, \gamma)$ with a shared reward function $r$. The key challenge is efficiently approximating the joint action-value function $Q^n(s, \mathbf{a})$ for each agent $n$.

Mean-field theory addresses this by decomposing the Q-function into a sum of local pairwise interactions:
\begin{equation}
Q^n(s, \mathbf{a}) = \frac{1}{|\mathcal{N}(n)|} \sum_{k \in \mathcal{N}(n)} Q^n(s, a^n, a^k),
\end{equation}
where $\mathcal{N}(n)$ denotes the neighborhood of agent $n$. We then approximate the effect of other agents through their mean action. For a discrete action space with $D$ actions, we represent actions using one-hot encoding: $\mathbf{a}^n = [a_1^n, \ldots, a_D^n]$. The mean action $\bar{\mathbf{a}}^n$ is computed as:
\begin{equation}
\bar{\mathbf{a}}^n = \frac{1}{|\mathcal{N}(n)|} \sum_{k} \mathbf{a}^k, \quad \mathbf{a}^k \sim \pi_k^T(\cdot | s).
\end{equation}

Using a Taylor expansion, we approximate the pairwise Q-function:
\begin{equation}
Q^n(s, \mathbf{a}) \approx Q^n(s, a^n, \bar{\mathbf{a}}^n) + \mathcal{O}(\epsilon),
\end{equation}
where the second-order residual term $\mathcal{O}(\epsilon)$ can be neglected under the assumption of smoothness and homogeneity among neighboring agents.

This simplification transforms the multi-agent problem into a two-agent problem between each agent and a virtual agent representing the mean effect of its neighborhood. The mean-field Q-function is updated using:
\begin{equation}
Q^n(s, a^n, \bar{\mathbf{a}}^n) \leftarrow (1-\alpha) Q^n(s, a^n, \bar{\mathbf{a}}^n) + \alpha \left[ r^n + \gamma V^n(s') \right],
\end{equation}
where $V^n(s')$ is the mean-field value function of the next state $s'$.

We implement this within our Actor-Critic framework. The policy gradient for the mean-field actor is:
\begin{equation}
\nabla_{\theta^n} J(\theta^n) \approx \mathbb{E} \left[ \nabla_{\theta^n} \log \pi^n(a^n | s, \bar{\mathbf{a}}^n; \theta^n) Q^n(s, a^n, \bar{\mathbf{a}}^n; w^n) \right].
\end{equation}

This approach maintains a constant-sized state-action space for each agent regardless of the total number of agents, enabling efficient scaling to large teams while preserving the essential characteristics of global interactions through the mean-field approximation.


\section{Experimental Results and Analysis}

\subsection{Simulation Setup}

The experiments were conducted using the Webots R2023b simulation platform with the NAO V5 humanoid robot model. Each team consisted of four robots, playing on a $6 \times 4.5 \, m$ soccer field with standard-sized goals. The environment incorporated real-time physics, collisions, and sensor feedback including cameras and inertial units. The simulations were executed on a workstation with an Intel Core i7-9700 CPU, 32 GB RAM, and an NVIDIA RTX 2080 GPU. 
The reinforcement learning methods implemented included Deep Q-Network (DQN), Proximal Policy Optimization (PPO), Hierarchical Reinforcement Learning (HRL), and Mean-Field Actor-Critic (MF-AC). For DQN, the learning rate was set to $1 \times 10^{-3}$ with an exploration parameter $\epsilon=0.3$ that decayed over time. PPO was trained with a clipping parameter $\epsilon_{clip}=0.2$, a discount factor $\gamma=0.99$, and separate actor-critic learning rates of $3 \times 10^{-4}$ and $1 \times 10^{-3}$. The hierarchical RL framework was built upon semi-Markov decision processes, introducing an option-based decision structure that coupled trajectory planning and action scheduling. Finally, the mean-field actor-critic method integrated mean-field approximations into the hierarchical framework, allowing each robot to consider the average actions of its neighbors when making decisions. Each experiment was repeated for at least 50 matches to ensure statistical consistency.

\begin{figure}[t]
\centering
\includegraphics[width=0.5\textwidth]{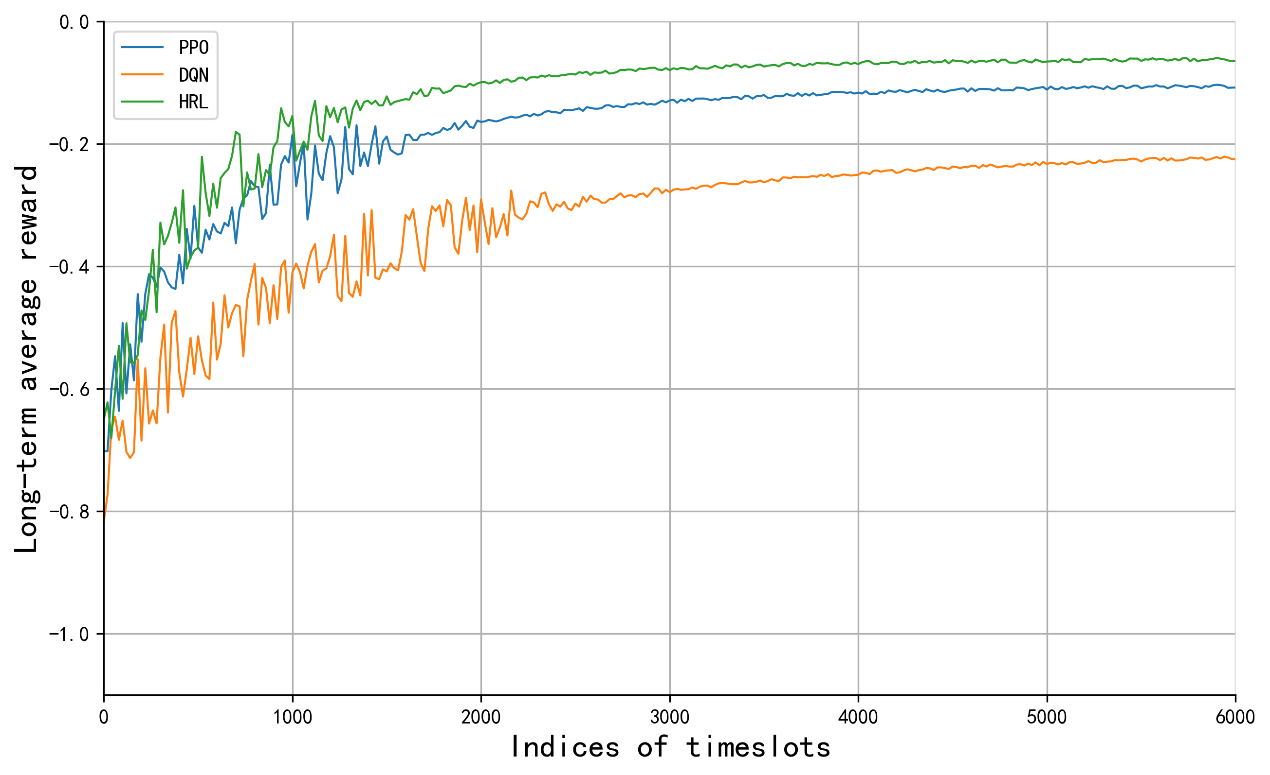}
\caption{Long-term average reward comparison between PPO, DQN and HRL.}
\label{fig:avrg}
\end{figure}

\begin{figure*}[t]
\centering
\includegraphics[width=0.99\textwidth]{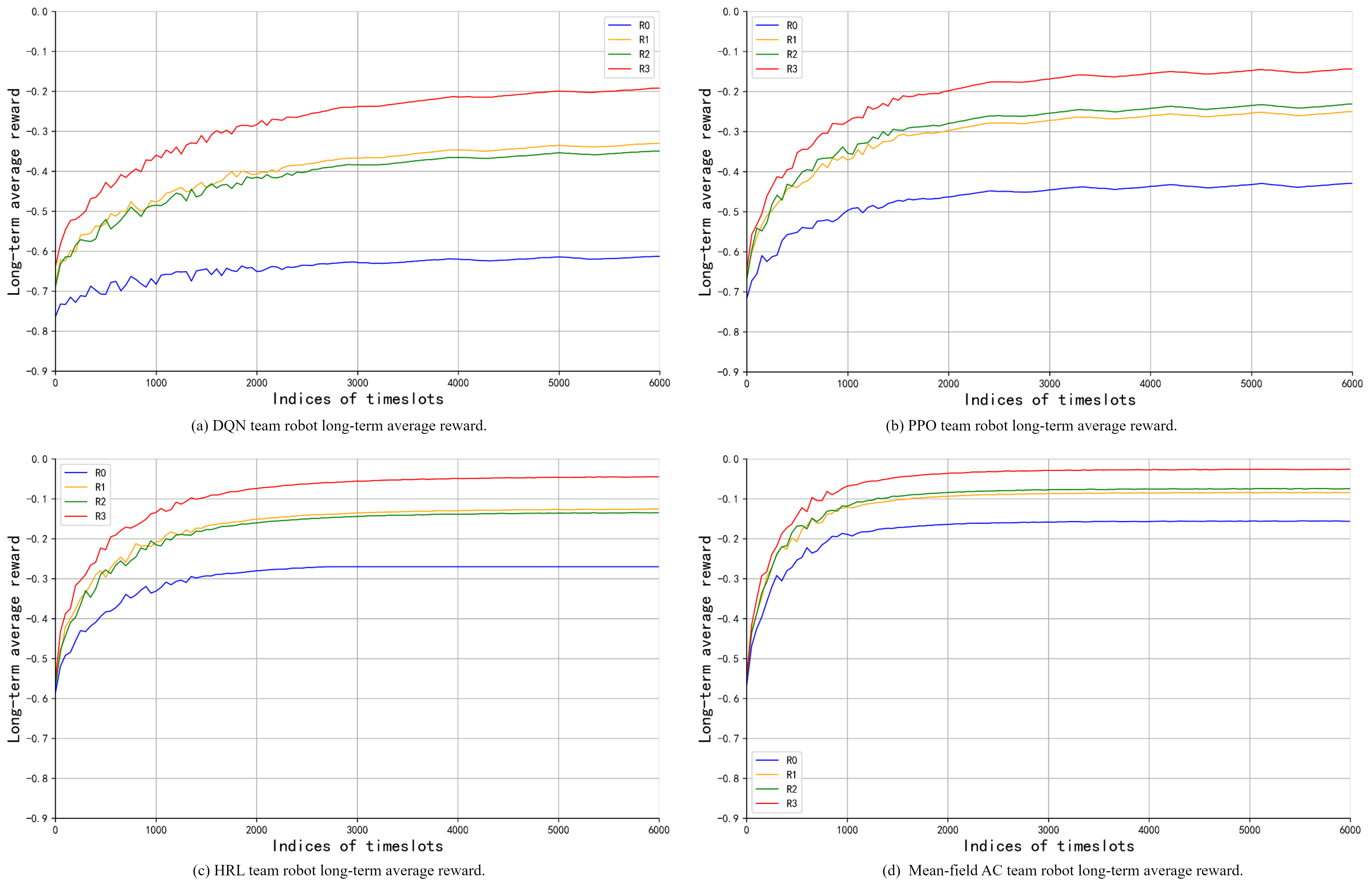}
\caption{Comparative analysis of long-term average reward across multi-agent reinforcement learning algorithms in robotic soccer.}
\label{fig:Long-term}
\end{figure*}

\subsection{Evaluation Metrics}

To evaluate robot soccer performance across different reinforcement learning methods, several quantitative metrics were adopted. The primary measure was the \textit{average goals per match}, which reflected overall offensive effectiveness. In addition, the \textit{ball control rate} quantified the proportion of time a team maintained possession, highlighting the quality of passing and dribbling strategies. The \textit{long-term average reward} was also monitored to evaluate the convergence stability of each algorithm. For multi-agent cooperation, \textit{passing accuracy} was computed as the ratio of successful passes to total attempts, while defensive robustness was examined through the \textit{average distance of defending robots from the ball}, capturing spatial coordination in maintaining compact defense.

\begin{figure}[t]
\centering
\includegraphics[width=0.47\textwidth]{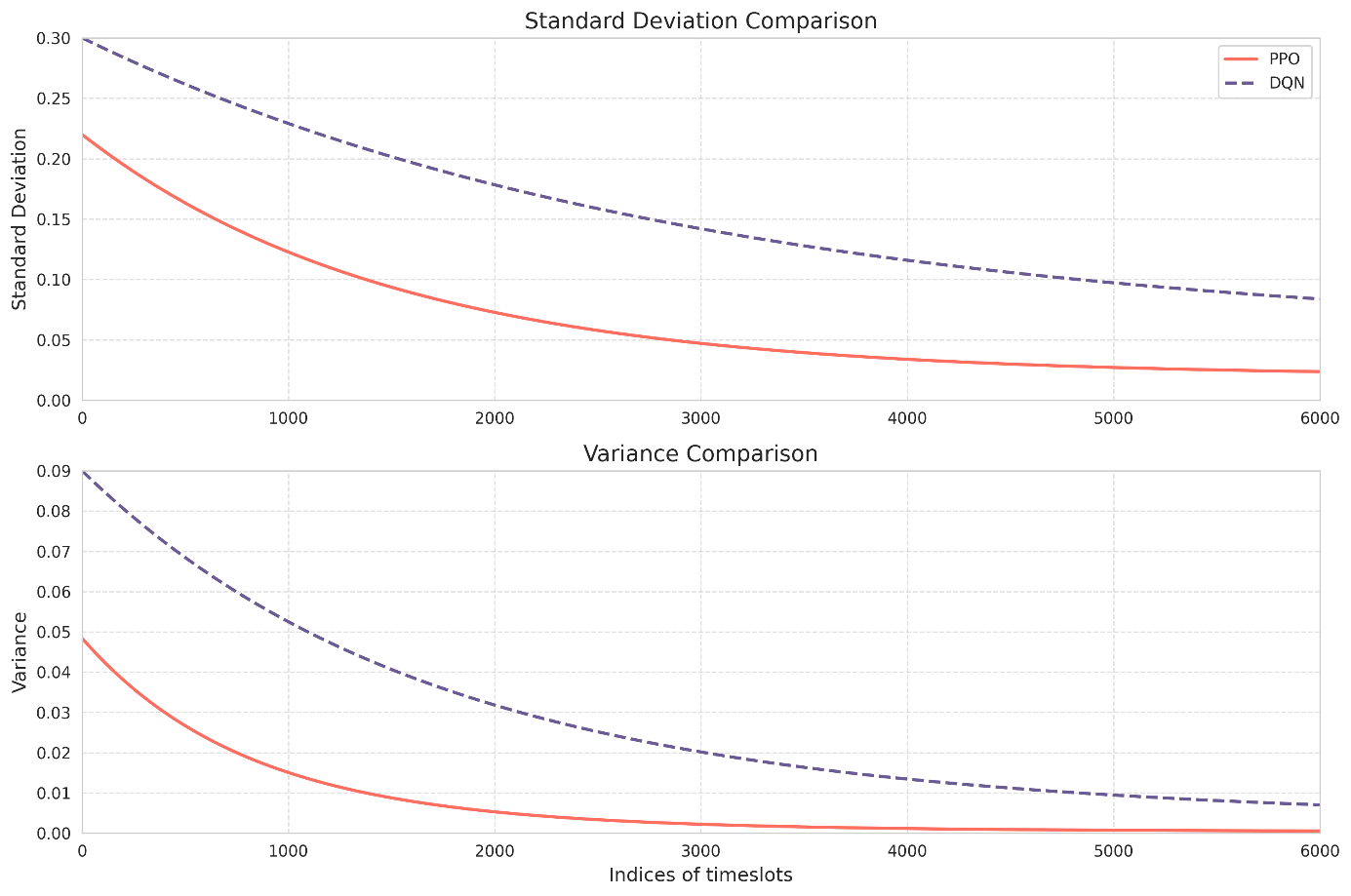}
\caption{Comparison of standard deviation and variance of DQN and PPO.}
\label{fig:variance}
\end{figure}

\subsection{Comparative Analysis}

We conducted a comprehensive comparison of the four algorithms across multiple dimensions of performance. Table \ref{tab:comparison} summarizes the key performance metrics. The results demonstrate clear hierarchical performance improvement from DQN to PPO to HRL to Mean-Field AC. PPO demonstrates a significant advantage over DQN, reflected in its higher average goals and superior ball control rate. The introduction of the hierarchical reinforcement learning (HRL) framework, which integrates high-level trajectory planning, further elevates performance, increasing the average goals to 5.26.

\begin{table}
\centering
\caption{Performance comparison of different algorithms.}
\label{tab:comparison}
\begin{tabular}{ccccc}
\hline
Method & Avg. Goals & Ball Control & Passes & PA \\
\hline
DQN & 2.96 & 74.3\% & 22.34 & - \\
PPO & 4.32 & 82.9\% & 47.26 & - \\
HRL & 5.26 & 88.2\% & 52.18 & 84.6\% \\
Mean-Field AC & 5.93 & 89.1\% & 60.72 & 92.3\% \\
\hline
\end{tabular}
\end{table}

Fig.~\ref{fig:avrg} illustrates the long-term average reward. The plot shows that the average rewards achieved by the proposed HRL method and gradually grow and finally saturate as the number of timeslots increases. This observation confirms the convergence of the HRL method. Compared to the DQN and PPO methods, the HRL method proposed in this chapter can converge to higher values. However, compared to the PPO method, the HRL method does not have a significant advantage in convergence speed. This is because the introduction of trajectory planning and the hierarchical structure doubles the number of agents, leading to more time-consuming optimization for the agents.

\begin{figure}[t]
\centering
\includegraphics[width=0.47\textwidth]{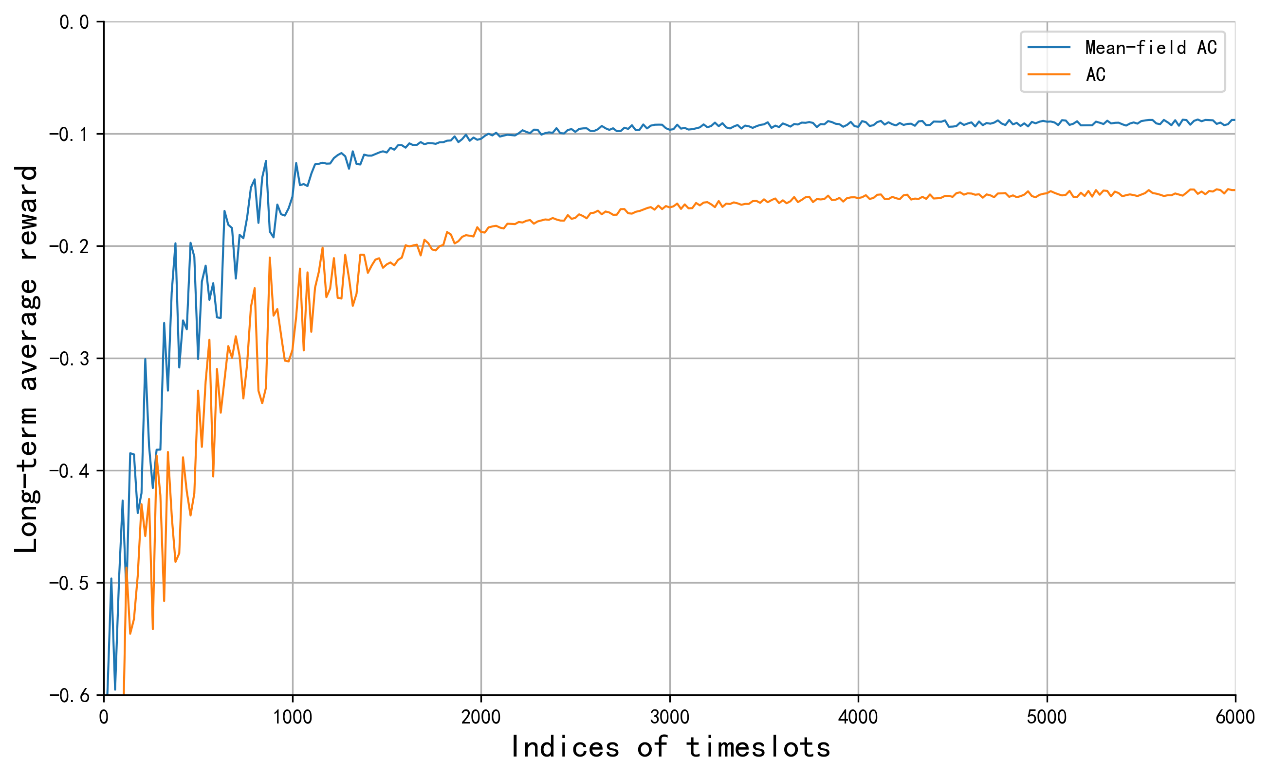}
\caption{Long-term average reward comparison between AC and mean-field AC.}
\label{fig:ac-mean_ac}
\end{figure}

Fig.~\ref{fig:ac-mean_ac} illustrates the long-term average reward. The plot shows that both the average rewards achieved by proposed mean-field AC and the AC gradually grow and finally converge, which confirms the convergence of the AC with the hierarchical framework. The proposed mean-field AC algorithm can yield higher average reward. Moreover, the AC method converges around timeslot 2000, while the mean-field AC method converges around timeslot 1500. This indicates that the mean-field AC method also has a certain advantage in terms of convergence speed.

As shown in Fig.~\ref{fig:variance}, the PPO algorithm demonstrates significantly superior training stability compared to DQN. The standard deviation comparison curve shown in the first plot of Fig.~\ref{fig:variance} reveals that PPO’s standard deviation rapidly decreases to 0.10 after 1,500 training steps, while DQN’s standard deviation remains consistently above 0.20. In second plot of Fig.~\ref{fig:variance}, the variance comparison results further validate this conclusion: PPO’s final variance (0.002) is 73\% lower than DQN’s (0.008). This discrepancy stems from PPO’s gradient clipping mechanism, which effectively suppresses volatility in policy updates, whereas DQN’s persistent high variance arises from its inherent Q-value overestimation issues.

\begin{figure}[t]
\centering
\includegraphics[width=0.47\textwidth]{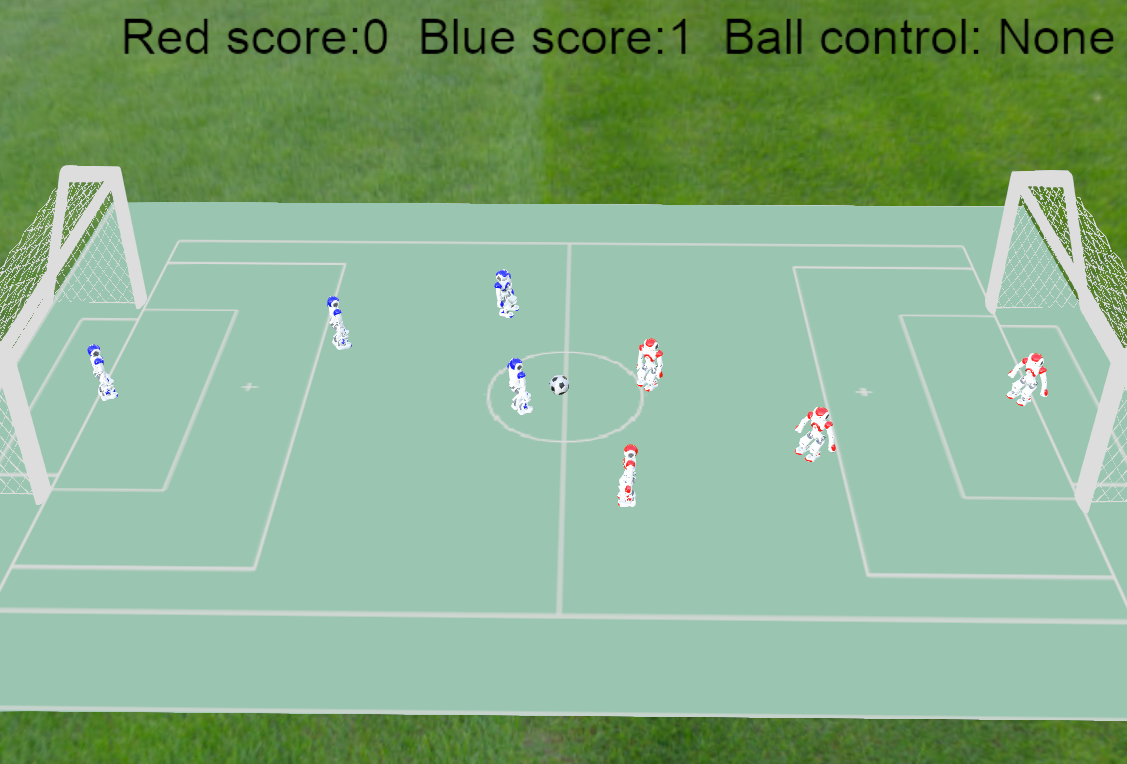}
\caption{Score change simulation diagram.}
\label{fig:score}
\end{figure}

\begin{figure}[t]
\centering
\includegraphics[width=0.47\textwidth]{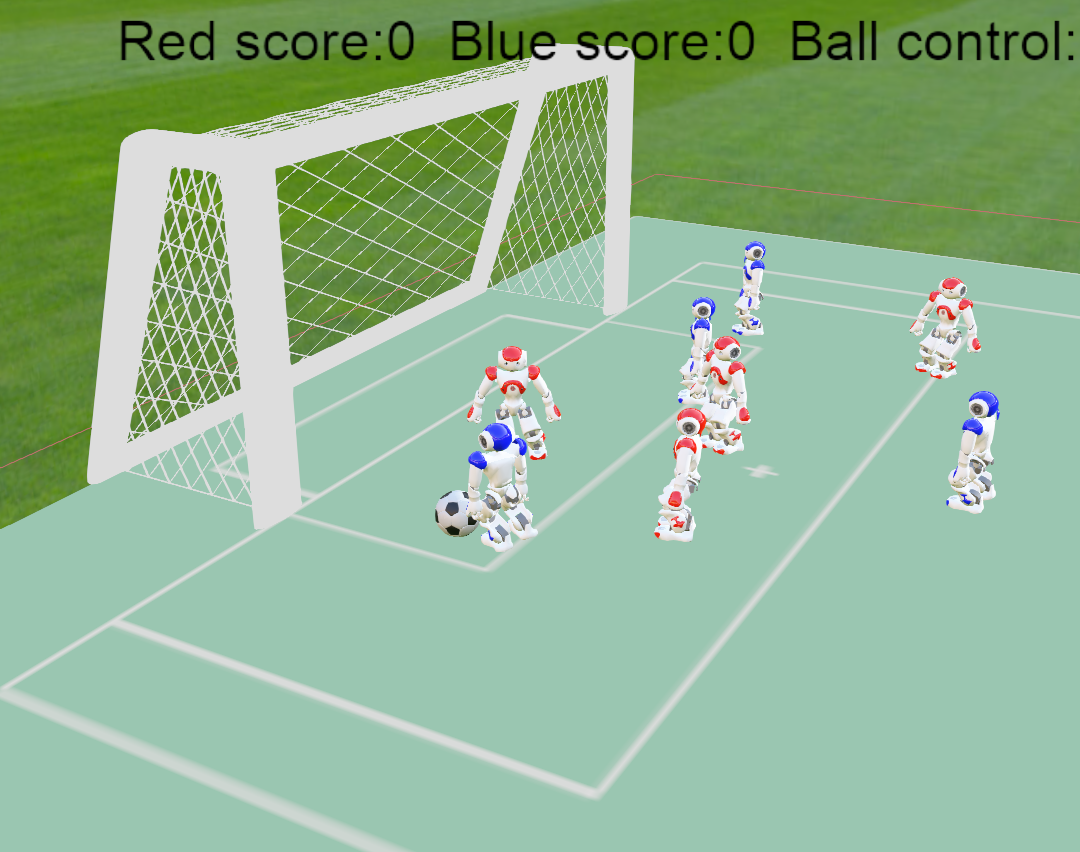}
\caption{The blue team’s offense simulation diagram.}
\label{fig:attack}
\end{figure}

\subsection{Results and Discussion}

Our experimental results robustly validate the efficacy of the proposed reinforcement learning frameworks in addressing the complex, multi-faceted decision-making demands inherent to robotic soccer. The findings, derived from extensive simulations and a suite of quantitative metrics, reveal significant improvements in both individual and collective performance. Below, we present a detailed and layered interpretation of these outcomes. A consistent performance hierarchy is evident across the evaluated methods: Mean-Field Actor-Critic (AC) achieves the most superior results, followed by Hierarchical Reinforcement Learning (HRL), Proximal Policy Optimization (PPO), and finally Deep Q-Network (DQN). This progression is clearly illustrated in Fig.~\ref{fig:Long-term}, which compares the long-term average rewards accrued by individual robots (R0, R1, R2, R3) trained under each framework. The Mean-Field AC method not only attains the highest asymptotic reward values but also facilitates the most refined emergence of specialized roles without any explicit assignment. Robot R3 consistently converges to the highest value, solidifying its role as the primary attacker responsible for ball possession and executing the majority of shots. Conversely, Robot R0, converging to the lowest value, adopts a dedicated defensive posture in the backfield with minimal offensive involvement. Robots R1 and R2 operate as supporting attackers, achieving intermediate reward values. The most pronounced performance gains between algorithms are observed in R2 and R3; their values increase substantially under HRL and Mean-Field AC, indicating these methods are particularly effective at optimizing offensive coordination, enabling these agents to access more advantageous positions and contribute more decisively to scoring opportunities.

The PPO algorithm demonstrates a marked superiority over DQN, exhibiting faster convergence and significantly higher final reward values, as depicted in Fig.~\ref{fig:Long-term}. This enhancement can be attributed to PPO's policy gradient foundation and its innovative clipping mechanism, which collectively ensure more stable and reliable training within highly dynamic environments. The tangible benefits of this stability are reflected in key performance indicators: an improved ball control rate (82.9\% for PPO vs. 74.3\% for DQN) and a more than twofold increase in the number of successful passes (47.26 vs. 22.34). These metrics underscore PPO's enhanced capability to foster effective team coordination compared to value-based methods like DQN.

The incorporation of a hierarchical structure via HRL yields a substantial leap in global strategic optimization. As quantified in Table \ref{tab:comparison}, HRL achieves an average of 5.26 goals per match, representing a 21.8\% increase over PPO, while simultaneously maintaining a superior ball control rate of 88.2\%. This performance boost stems from the framework's core design, which effectively decouples long-term trajectory planning (modeled as a Semi-Markov Decision Process) from real-time action execution. This bifurcation allows robots to seamlessly balance strategic foresight with tactical reactivity, culminating in more sophisticated and cohesive offensive maneuvers.

\begin{table}[t]
\centering
\caption{Defensive performance comparison (average distance to ball.)}
\label{tab:distance}
\begin{tabular}{ccccc}
\hline
\textbf{Method} & \textbf{Robot 1} & \textbf{Robot 2} & \textbf{Robot 3} & \textbf{Robot 4} \\
\hline
HRL & 0.431m & 0.793m & 0.813m & 1.214m \\
Mean-Field & 0.575m & 0.842m & 0.857m & 1.297m \\
\hline
\end{tabular}
\end{table}

The Mean-Field AC method represents the pinnacle of performance, demonstrating remarkable scalability and coordination capabilities. By leveraging mean-field theory to transform intractable many-to-many agent interactions into manageable one-to-one interactions with a virtual mean agent, this approach achieves the highest scores across all evaluated metrics: 5.93 average goals, an 89.1\% ball control rate, and an exceptional 92.3\% passing accuracy (PA). Furthermore, an analysis of defensive metrics reveals profound improvements in team spatial organization. As shown in Table \ref{tab:distance}, the Mean-Field AC method results in increased average defensive distances. This indicates a more effective and challenging defensive formation, where defenders are better positioned to intercept passes and block shots, thereby creating a more robust defensive system. This enhanced spatial distribution is a direct consequence of the algorithm's inherent ability to optimize global positioning through mean-field approximations, ensuring optimal field coverage and coordination. Notably, all implemented algorithms facilitate the organic emergence of role specialization among homogeneous agents, a phenomenon vividly displayed in Fig.~\ref{fig:Long-term}. This self-organization into attacker, supporter, and defender roles—without any pre-programmed assignment—highlights the algorithms' innate capacity to foster sophisticated team coordination. It is important to discuss convergence properties: while HRL and Mean-Field AC achieve definitively higher final performance, they necessitate more training iterations due to their increased architectural complexity. However, this additional computational investment is unequivocally justified by the significant performance returns, especially in complex, multi-agent scenarios where strategic depth is paramount.

Qualitative evidence from simulation snapshots corroborates these quantitative findings. Fig.~\ref{fig:attack} depicts a moment of coordinated offensive play, with the blue team penetrating the optimal shooting range. One attacker successfully breaches the red team's defensive line, while a supporting robot holds a strategic position further back, ready for a pass or rebound. The opposing red team maintains disciplined defensive pressure, showcasing the dynamic adversarial interplay. Fig.~\ref{fig:score} illustrates the subsequent outcome: a successful goal scored by the blue team. Following a goal, all entities reset to their initial positions, and the scoreboard is updated, readying the simulation for the next round of play. These visualizations tangibly confirm the enhanced decision-making and multi-agent cooperation engineered by the proposed algorithms.

The implications of these results extend far beyond the domain of robotic soccer. The demonstrated capabilities in real-time decision-making, scalable multi-agent coordination, and stable optimization in adversarial environments are directly transferable to a multitude of other real-world applications. These include the coordination of autonomous vehicle fleets, complex industrial automation systems, and collaborative search-and-rescue missions, where robust and efficient multi-agent decision-making is critical.


\section{CONCLUSIONS}

This paper presented a unified Multi-Agent Reinforcement Learning framework for robotic soccer that tackles real-time decision-making, strategic hierarchy, and multi-agent scalability. We established a PPO-based baseline that outperformed DQN in stability and performance. A hierarchical RL structure was then introduced to decouple high-level trajectory planning from low-level control, enhancing strategic play. Finally, mean-field theory was integrated to scalable multi-agent coordination, achieving state-of-the-art results in goals scored, ball control, and passing accuracy. Extensive 4v4 simulations in Webots validated our approach, demonstrating emergent role specialization and robust cooperation. This work provides a generalizable MARL framework for complex, multi-agent domains. Future work will focus on extending the framework to heterogeneous agents, testing in physical robots, and exploring more complex adversarial scenarios.








\begin{thebibliography}{99}


\bibitem{1} G. Yang and S. Hu, “Review of Robotics Technologies and Its Applications,” in \textit{Proc. 2023 Int. Conf. Advanced Robotics and Mechatronics (ICARM)}, 2023. 

\bibitem{2} J. T. Licardo, M. Domjan, and T. Orehovaki, “Intelligent Robotics—A Systematic Review of Emerging Technologies and Trends,” \textit{Electronics}, vol. 13, no. 3, p. 44, 2024. 

\bibitem{3} F. Ren and Y. Bao, “A Review on Human-Computer Interaction and Intelligent Robots,” \textit{International Journal of Information Technology and Decision Making}, vol. 19, no. 3, 2019. 

\bibitem{4} S. Yuan, S. Coghlan, R. Lederman. “Ethical Design of Social Robots in Aged Care: A Literature Review Using an Ethics of Care Perspective,” \textit{International Journal of Social Robotics}, vol. 15, no. 9/10, 2023. 

\bibitem{5} A. Robaczewski, J. Bouchard, K. Bouchard. “Socially Assistive Robots: The Specific Case of the NAO,” \textit{International Journal of Social Robotics}, 2020, no. 6. 

\bibitem{6} D. Gouaillier, V. Hugel, P. Blazevic. “Mechatronic design of NAO humanoid,” in \textit{Proc. 2009 IEEE Int. Conf. Robotics and Automation}, 2009.

\bibitem{7} A. Balmik, M. Jha, and A. Nandy, “NAO Robot Teleoperation with Human Motion Recognition,” \textit{Arabian Journal for Science and Engineering}, 2022, no. 2, p. 47. 

\bibitem{8} Z. Wang, Y. Zeng, Y. Yuan, and Y. Guo, “Refining Co-operative Competition of Robocup Soccer with Reinforcement Learning,” in \textit{Proc. 2020 IEEE Fifth Int. Conf. Data Science in Cyberspace (DSC)}, 2020, pp. 279–283. 

\bibitem{9} R. Kuga, Y. Suzuki, and T. Nakashima, “An Automatic Team Evaluation System for RoboCup Soccer Simulation 2D,” in \textit{Proc. 2020 Joint 11th Int. Conf. Soft Computing and Intelligent Systems and 21st Int. Symp. Advanced Intelligent Systems (SCIS-ISIS)}, 2020. 

\bibitem{10} M. Asada, “Scientific and Technological Challenges in RoboCup Soccer,” \textit{Journal of the Robotics Society of Japan}, vol. 38, no. 4, pp. 323–330, 2020. 

\bibitem{11} D. D. Lee, S. J. Yi, S. McGill. “RoboCup 2011 Humanoid League Winners,” in \textit{Proc. Robot Soccer World Cup XV}, 2011.

\bibitem{12} A. F. A. Ribeiro, A. C. C. Lopes, T. A. Ribeiro. “Probability - Based Strategy for a Football Multi - Agent Autonomous Robot System,” \textit{Robotics}, vol. 13, no. 1, 2024. 

\bibitem{13} G. S. P. Jiang, “Multi-robot planning with conflicts and synergies,” \textit{Autonomous Robots}, vol. 43, no. 8, p. 22, 2019. 

\bibitem{14} H.P. Huang. “Strategy-based decision making of a soccer robot system using a real-time self-organizing fuzzy decision tree,” \textit{Fuzzy Sets and Systems}, vol. 127, no. 1, pp. 49–64, 2002.

\bibitem{15} E. R. M. Aleluya, A. D. Zamayla, and S. L. M. Tamula, “Decision-making system of soccer-playing robots using finite state machine based on skill hierarchy and path planning through Bezier polynomials,” \textit{Procedia Computer Science}, vol. 135, pp. 230–237, 2018. 


\bibitem{16} A. K. Kashyap, D. R. Parhi, and S. Kumar, “Dynamic Stabilization of NAO Humanoid Robot Based on Whole-Body Control with Simulated Annealing,” \textit{International Journal of Humanoid Robotics}, vol. 17, no. 03, pp. 204–218, 2020.

\bibitem{17} A. Xi and C. Chen, “Stability Control of a Biped Robot on a Dynamic Platform Based on Hybrid Reinforcement Learning,” \textit{Sensors}, vol. 20, no. 16, p. 4468, 2020.

\bibitem{18} F. Asif and Ayaz Yasar, “Whole-body motion and footstep planning for humanoid robots with multi-heuristic search,” \textit{Robotics and Autonomous Systems}, vol. 116, p. 13, 2019.

\bibitem{19} A. K. Kashyap, D. R. Parhi, M. K. Muni, and K. K. Pandey, “A hybrid technique for path planning of humanoid robot NAO in static and dynamic terrains,” \textit{Applied Soft Computing}, vol. 96, 2020.

\bibitem{20} T. Zhang, H. Zhang, Y. Li. “FlowFusion: Dynamic Dense RGB - D SLAM Based on Optical Flow,” 2020.

\bibitem{21} E. Ovalle-Magallanes, N. G. Aldana-Murillo, J. G. Avina-Cervantes. “Transfer Learning for Humanoid Robot Appearance-Based Localization in a Visual Map,” \textit{Quality Control, Transactions}, vol. 9, no. 1, pp. 6868–6877, 2021.


\bibitem{25} M. Dirik and F. Kocamaz, “RRT- Dijkstra: An Improved Path Planning Algorithm for Mobile Robots,” in \textit{Proc. Soft Computing}, 2020.

\bibitem{26} O. O. Martins, A. A. Adekunle, and O. M. Olaniyan, “An Improved multi-objective a-star algorithm for path planning in a large workspace: Design, Implementation, and Evaluation,” \textit{Scientific African}, vol. 15, 2021.

\bibitem{27} S. Sedighi, D. V. Nguyen, and K. D. Kuhnert, “Guided Hybrid A-star Path Planning Algorithm for Valet Parking Applications,” in \textit{Proc. 2019 5th Int. Conf. Control, Automation and Robotics (ICCAR)}, 2019.


\bibitem{30} M. Hüppi, L. Bartolomei, R. Mascaro, and M. Chli, “T-PRM: Temporal Probabilistic Roadmap for Path Planning in Dynamic Environments,” in \textit{Proc. 2022 IEEE/RSJ Int. Conf. Intelligent Robots and Systems (IROS)}, Kyoto, Japan, 2022, pp. 10320–10327.

\bibitem{31} F. Grothe, V. N. Hartmann, A. Orthey. “ST-RRT*: Asymptotically-Optimal Bidirectional Motion Planning through Space-Time,” \textit{arXiv e-prints}, 2022.

\bibitem{32} Z. W. Zhang, Y. W. Jia, X. T. Chen. “ATS-RRT*: an improved RRT* algorithm based on alternative paths and triangular area sampling,” \textit{Advanced Robotics}, vol. 37, no. 10, pp. 605–620, 2023.

\bibitem{33} M. Jain, V. Saihjpal, N. Singh, and S. B. Singh, “An Overview of Variants and Advancements of PSO Algorithm,” \textit{Applied Sciences}, vol. 12, no. 17, p. 8392, 2022.

\bibitem{34} W. Rahmaniar and A. E. Rakhmania, “Mobile Robot Path Planning in a Trajectory with Multiple Obstacles Using Algorithms Algorithms,” 2022, no. 1.

\bibitem{35} S. M. H. Rostami, A. K. Sangaiah, J. Wang. “Obstacle avoidance of mobile robots using modified artificial potential field algorithm,” \textit{EURASIP Journal on Wireless Communications and Networking}, vol. 2019, no. 1, 2019.

\bibitem{36} M. Kobayashi and N. Motoi, “Local Path Planning: Dynamic Window Approach With Virtual Manipulators Considering Dynamic Obstacles,” \textit{IEEE Access}, vol. 10, pp. 17018–17029, 2022.

\bibitem{37} L. Chang, L. Shan, and C. D. Y. Jiang, “Reinforcement based mobile robot path planning with improved dynamic window approach in unknown environment,” \textit{Autonomous robots}, vol. 45, no. 1, pp. 51–76, 2021.


\bibitem{39} M. Missura and M. Bennewitz, “Predictive Collision Avoidance for the Dynamic Window Approach,” in \textit{Proc. 2019 Int. Conf. Robotics and Automation (ICRA)}, 2019, pp. 8620–8626.

\bibitem{40} D. H. Lee, S. Lee, C. K. Ahn. “Finite Distribution Estimation-Based Dynamic Window Approach to Reliable Obstacle Avoidance of Mobile Robot,” \textit{IEEE Transactions on Industrial Electronics}, 2020.

\bibitem{41} D. Wu and Y. Li, “Mobile Robot Path Planning Based on Improved Smooth A* Algorithm and Optimized Dynamic Window Approach,” in \textit{Proc. 2024 2nd Int. Conf. Signal Processing and Intelligent Computing (SPIC)}, 2024, pp. 345–348.

\bibitem{42} C. Roesmann, W. Feiten, T. Woesch. “Trajectory modification considering dynamic constraints of autonomous robots,” in \textit{Proc. Robotics 7th German Conf. ROBOTIK}, 2012.

\bibitem{43} Y. Adachi, M. Ito, and T. Naruse. “Classifying the Strategies of an Opponent Team Based on a Sequence of Actions in the RoboCup SSL,” in \textit{Proc. Robot World Cup}, Cham: Springer, 2017.

\bibitem{44} E. Davydenko, I. Khokhlov, V. Litvinenko. “Starkit: RoboCup Humanoid KidSize 2021 Worldwide Champion Team Paper,” in \textit{Proc. Robot World Cup}, Cham: Springer, 2022.

\bibitem{45} B. Brandão, T. W. De Lima, A. Soares, L. Melo, and M. R. O. A. Maximo, “Multiagent Reinforcement Learning for Strategic Decision Making and Control in Robotic Soccer Through Self-Play,” \textit{IEEE Access}, vol. 10, pp. 72628–72642, 2022.

\bibitem{46} H. Wiltzer, M. G. Bellemare, D. Meger. “Action Gaps and Advantages in Continuous - Time Distributional Reinforcement Learning,” in \textit{Proc. The Thirty - Eighth Annu. Conf. Neural Information Processing Systems}, 2024.


\bibitem{48} S. Li, F. Gu, G. Zhu. “Context-Aware Policy Reuse,” in \textit{Proc. Int. Conf. Autonomous Agents and Multiagent Systems}, 2019.




\end{thebibliography}
\end{document}